\documentclass[runningheads]{llncs}

\usepackage{graphicx}
\graphicspath{{figures/}}
\usepackage{booktabs}
\usepackage{amsmath}
\usepackage{siunitx}
\usepackage[hidelinks]{hyperref}

\newcommand\blfootnote[1]{%
  \begingroup\renewcommand\thefootnote{}\footnote{#1}\addtocounter{footnote}{-1}\endgroup}

\begin{document}

\title{Intelligent Character Recognition of Handwritten Forms\\
with Deep Neural Networks}
\titlerunning{Intelligent Character Recognition of Handwritten Forms}

\author{Hartwig Grabowski\inst{1}}
\authorrunning{H. Grabowski}
\institute{Institute for Machine Learning and Analytics (IMLA)\\
Offenburg University, Germany\\
\email{hartwig.grabowski@hs-offenburg.de}}

\maketitle

\blfootnote{\textbf{Preprint / author's accepted manuscript.} This is the
author's version of a chapter published in \emph{Towards AI-Aided Invention
and Innovation}, IFIP Advances in Information and Communication Technology,
Springer Nature Switzerland, 2023, pp.~81--94. The final authenticated version
is available online at \url{https://doi.org/10.1007/978-3-031-42532-5_6} and is
the definitive version of record. Please cite the published version.}

\begin{abstract}
The automatic processing of handwritten forms remains a challenging task,
wherein detection and subsequent classification of handwritten characters are
essential steps. We describe a novel approach, in which both steps --
detection and classification -- are executed in one task through a deep neural
network. Therefore, training data is not annotated by hand, but manufactured
artificially from the underlying forms and yet existing datasets. It can be
demonstrated that this single-task approach is superior in comparison to the
state-of-the-art two-task approach. The current study focuses on hand-written
Latin letters and employs the EMNIST data set. However, limitations were
identified with this data set, necessitating further customization. Finally,
an overall recognition rate of \SI{88.28}{\percent} was attained on real data
obtained from a written exam.
\end{abstract}

\section{Introduction}
Despite the prevalence of digital technology, paper forms continue to be used
in various contexts, often due to legal requirements or user preferences. Many
applications still rely on paper forms that are filled out by hand, such as
surveys, job applications, medical records or government forms. Automating the
processing of these forms can save time and reduce errors compared to manual
data entry~\cite{r1,r2}.

\begin{figure}[t]
\centering
\includegraphics[width=0.85\textwidth]{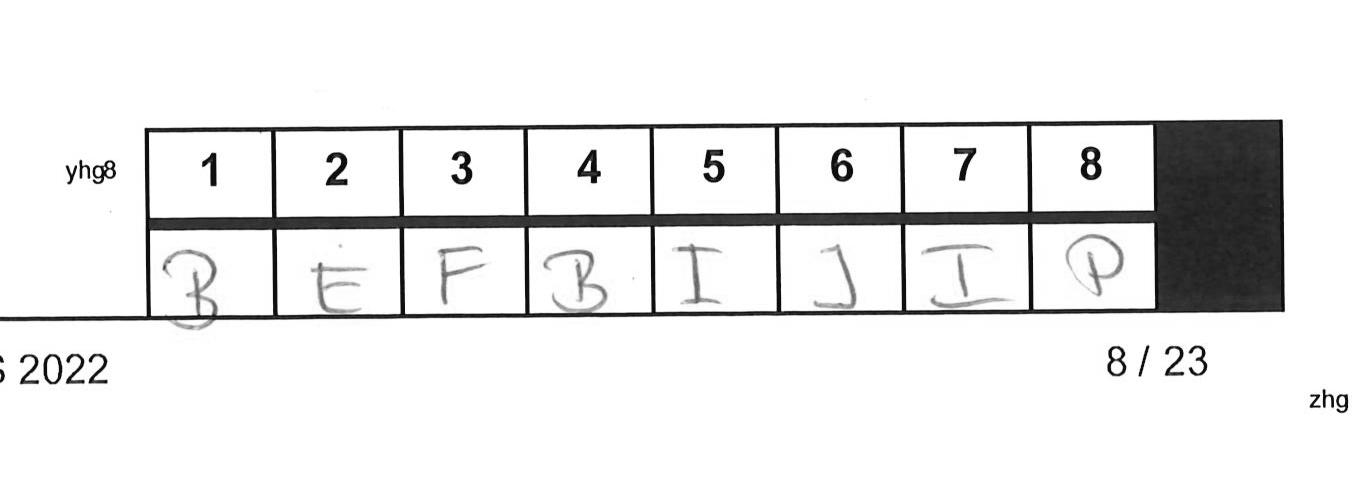}
\caption{A table is printed on the lower right side of the paper. Each column
represents one question. Each question must be answered by a capital Latin
letter.}
\label{fig:table}
\end{figure}

Other applications for automated form processing is the processing of health
records~\cite{r3}, the processing of bank cheques~\cite{r4} or the processing
of job application forms~\cite{r5}. Our use-case is the processing of printed
exams, where the possible answers are given by capital Latin letters
(``A'', ``B'', \ldots, ``Z''), which must be entered into a table. For further
processing, the letters in the table must be recognized by our system
(see Fig.~\ref{fig:table}).

Automated form processing is strongly related with Character Recognition,
which can be categorized into (a) the recognition of machine printed
characters, which is substituted by the term Optical Character Recognition
(OCR) and (b) Handwritten Character Recognition (HCR). HCR can further be
divided into online- and offline systems. In general, online systems use the
runtime input (e.g.\ the trajectory of the input pen), whereas offline systems
are based on pixel images, often obtained by scanning a printed
page~\cite{r6,r7,r8,r9}.

Whereas HCR focuses more on the classification part of the cropped image
itself, Intelligent Character Recognition (ICR) is a broader term that
addresses the character recognition task in different conditions, e.g.\ in
forms, on plates, in pictures (called ``text in the wild'')~\cite{r10,r11}.

\begin{figure}[t]
\centering
\includegraphics[width=\textwidth]{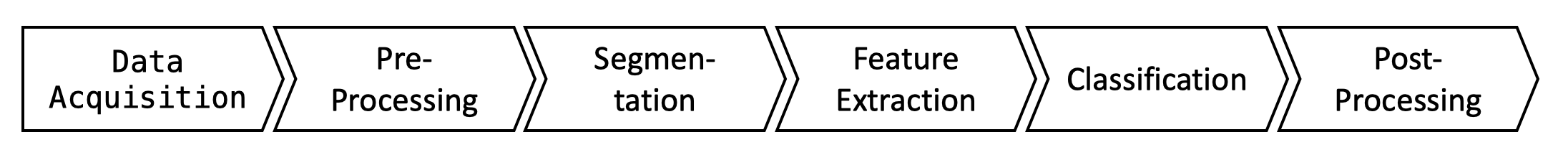}
\caption{The six processing steps for Intelligent Character Recognition (ICR).}
\label{fig:steps}
\end{figure}

Common approaches of HCR in forms and ICR involve six different processing
steps (see Fig.~\ref{fig:steps}): (1) data acquisition, (2) preprocessing,
(3) segmentation, (4) feature extraction, (5) classification and
(6) post-processing~\cite{r2,r6,r7,r12,r13,r14,r15,r16}. The next section
summarizes the literature concerning each of the six processing steps.

\section{Related Work}
\begin{enumerate}
\item \textbf{Data Acquisition:} The form to be processed is scanned or
photographed. The result is obtained as a bitmap image of formats like JPEG,
PNG etc.~\cite{r9,r17}. (Whether lossless compression is necessary depends on
the further process steps.)
\item \textbf{Preprocessing:} Various techniques, such as binarization,
complement, size normalization, morphological operation, noise removal using
filters, thinning, cleaning techniques, filtering mechanisms, thresholding, and
skeletonizing techniques are applied to eliminate undesirable elements from the
image~\cite{r18,r19}.
\item \textbf{Segmentation:} The preprocessed image is decomposed into
sub-images containing the individual characters. If tables are used, parallel
lines and the horizontal and vertical space can be used as features to extract
the table regions~\cite{r20,r21,r22,r23}. As of 2014, the ``Regions with
Convolutional Neural Network (CNN) features'' method (R-CNN) has been
developed~\cite{r24}, which locate single or multiple targets in digital
images. This approach has been evolved towards Fast-R-CNN~\cite{r25} and
Faster-R-CNN~\cite{r26} models, which are also called target detection
models~\cite{r27}. These models can also be used to detect the position and the
structure of tables in image-based documents~\cite{r28,r29}. Meanwhile more
complex models like YOLOv5, YOLOv7~\cite{r30} or Detectron2~\cite{r31} became
popular for detecting targets in images, e.g.\ a fine-tuned YOLOv5 network has
been trained for License Plate Recognition~\cite{r32,r33} or for detecting Sui
script characters in endangered archives~\cite{r34}.
\item \textbf{Feature Extraction:} From the decomposed sub-image a vector of
features is derived. These features can be categorized in Global-Features
(Fourier Transforms, Wavelets, Hough Transforms, etc.), Statistical Features
(Zoning, Crossing and Distances, etc.) and Geometrical Features (Strokes, Chain
Codes, etc.)~\cite{r7,r35}. However, these features are strongly related to the
underlying classification method and with the upcoming of the CNN and its
variants, feature extraction is applied implicitly by the CNN itself in the
convolution layers.
\item \textbf{Classification:} The extracted features are mapped to a specific
character. There are various techniques used for classification, which can be
categorized into Artificial Neural Networks (ANN), Kernel Methods like Support
Vector Machines (SVMs), Statistical Methods, Template Matching Techniques and
Structural Pattern Recognition~\cite{r11,r14,r35}. However, in recent times
ANNs became more popular and they outperform SVMs: the current 8 best image
classifiers for EMNIST-Letterset~\cite{r36} all are ANNs~\cite{r37}, achieving
an accuracy from \SI{95.96}{\percent} to \SI{93.65}{\percent}~\cite{r38,r39,r40,r41,r42}.
Even less complex ANN-Models consisting of only 6 CNN-Layers and 2 Dense-Layers
combined with Batch-Normalization- and Max-Pooling-Layers achieve a
\SI{90.59}{\percent} accuracy on the EMNIST-Balanced Letterset~\cite{r43}.
\item \textbf{Post-processing:} After character classification, several
techniques can be implemented to enhance HCR accuracy. One of these techniques
involves using multiple classifiers to classify the image, which can be
utilized in parallel, cascading, or hierarchical ways~\cite{r44}. To further
refine HCR results, contextual analysis may be conducted by considering the
document and geometrical context of the image to minimize the risk of errors.
\end{enumerate}

\section{Datasets}
The approach presented here uses ANN for the classification task. However, for
training the ANN much training data is required and even specialized data
augmentation methods which use trained decoder networks to generate variations
of the sample characters require 200 and more samples for each
class~\cite{r40}. Therefore, it is advantageous to be able to use existing
datasets. For Latin letters, the EMNIST-Letterset~\cite{r36,r45} became popular
and is also used by our approach. There exist different subsets of the
EMNIST-Letterset, but the ``EMNIST Balanced Letter'' dataset consisting of 3000
samples (shape of 28x28 pixels) for each of 47 classes fits best to our
classification problem. The 47 classes represent the characters
`0123456789ABCDEFGHIJKLMNOPQRSTUVWXYZabdefghnqrt'. Samples of the lowercase
letters `ijklmopsuvwxyz' were merged with the class of their corresponding
uppercase letters. However, in our approach only characters A--Z are to be
detected.

\begin{figure}[t]
\centering
\includegraphics[width=\textwidth]{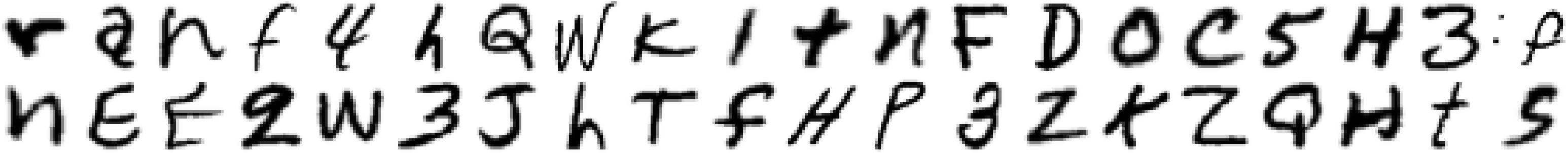}
\caption{The first 40 characters from the EMNIST Balanced Letterset. Each
sample has a shape of 28x28 pixels.}
\label{fig:emnist}
\end{figure}

\section{Initial Approach}
\subsection{Architecture}
\begin{figure}[t]
\centering
\includegraphics[width=0.85\textwidth]{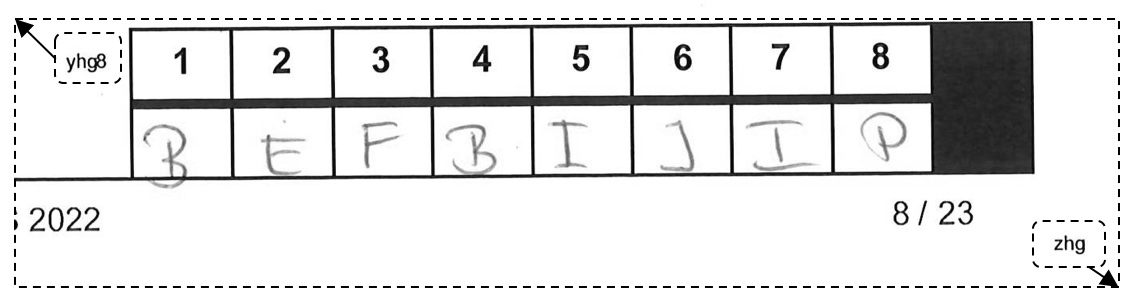}
\caption{Segmentation of the table is based on text marker detected with
tesseract~\cite{r46}.}
\label{fig:markers}
\end{figure}

\begin{figure}[t]
\centering
\includegraphics[width=0.7\textwidth]{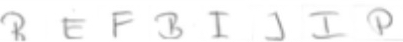}
\caption{Segmentation of each letter by cropping the hardcoded the positions of
the cells.}
\label{fig:segletters}
\end{figure}

\begin{figure}[t]
\centering
\includegraphics[width=\textwidth]{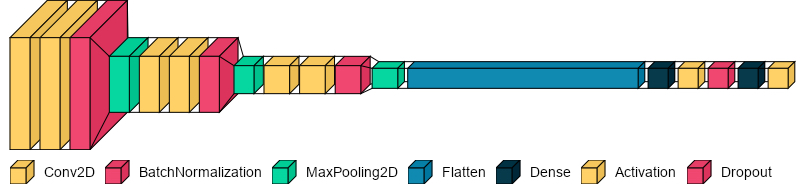}
\caption{The architecture of the CNN: Every two convolution layers are followed
by BatchNormalization and MaxPooling, Softmax is used for Output Layer.
Detailed description in~\cite{r43}.}
\label{fig:cnn}
\end{figure}

\begin{figure}[t]
\centering
\includegraphics[width=0.85\textwidth]{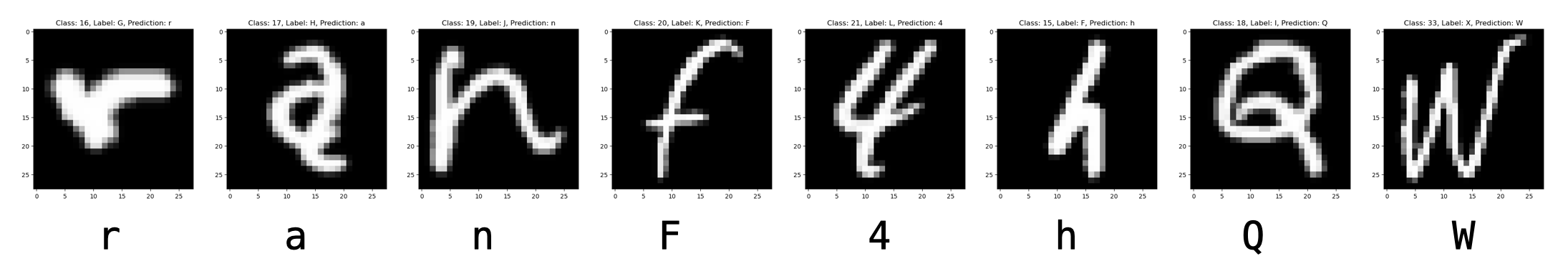}
\caption{The first 8 letters of the EMNIST dataset and their predicted
character by the CNN. All predictions are correct, letter `f' and `F' are
accumulated in one class.}
\label{fig:emnistpred}
\end{figure}

Our initial approach for recognizing the handwritten letters of the exam shown
in Fig.~\ref{fig:table} was based on the six previously described steps as
follows:
\begin{enumerate}
\item \textbf{Data-Acquisition} is done with the help of a document scanner,
which generates 300dpi image of the page in PDF format. If the page contains
grey-scaled colors only (e.g.\ the student writes with a black pencil), the
scanner automatically binarizes the image whereas colored pages (e.g.\ the
student writes with a blue pencil) are stored with the according RGB
information.
\item \textbf{Pre-processing:} Our form is provided by a table consisting of
two rows and up to nine columns. The table is always printed at the lower right
edge of the paper and for easy detection, we added two unique meaningless
abbreviation ``yhg'' and ``zhg'' at the upper left and lower right side of the
table (see Fig.~\ref{fig:markers}) serving as markers. With the help of
existing OCR application like tesseract~\cite{r46}, these markers and their
position can be detected and the table can be cropped from the whole page at the
markers position. Then, all cropped images are converted into gray-scale images
and stored without compression as images in PNG format.
\item \textbf{Segmentation:} As the position of the table is known due to the
markers and the shape of the table is well defined, the position of each cell
can be calculated and its content can be cropped. Each cropped image of each
table cell is then resized to 28x28 pixels (according to the EMNIST dataset)
and stored as a PNG image (see Fig.~\ref{fig:segletters}).
\item \textbf{Feature Extraction and Classification} are both performed in one
single task by a CNN similar to the one presented in~\cite{r43}. In order to be
comparable with existing approaches, we used all 47 classes although we only
need to recognize the 27 classes A--Z. The CNN was trained on the EMNIST
Balanced Letter Dataset consisting of 3000 samples (shape of 28x28 pixels) for
each of the 47 classes. 2600 samples were used for training and for validation
(thereof \SI{80}{\percent} training, \SI{20}{\percent} validation). 400 samples
were reserved for testing. An accuracy of \SI{89.23}{\percent} for the
validation and \SI{89.02}{\percent} for the testing-dataset was achieved. (Our
accuracy is below the \SI{90.56}{\percent} presented by~\cite{r43}, however no
pre-processing was executed and the CNN was adapted to 28x28 image size.) See
Fig.~\ref{fig:cnn}.
\item \textbf{Post-Processing:} The classified letter with the maximal
probability is chosen and no further post-processing is executed.
Fig.~\ref{fig:emnistpred} shows the result applied to the first eight samples
of the EMNIST data set.
\end{enumerate}

\subsection{Results}
\begin{figure}[t]
\centering
\includegraphics[width=0.85\textwidth]{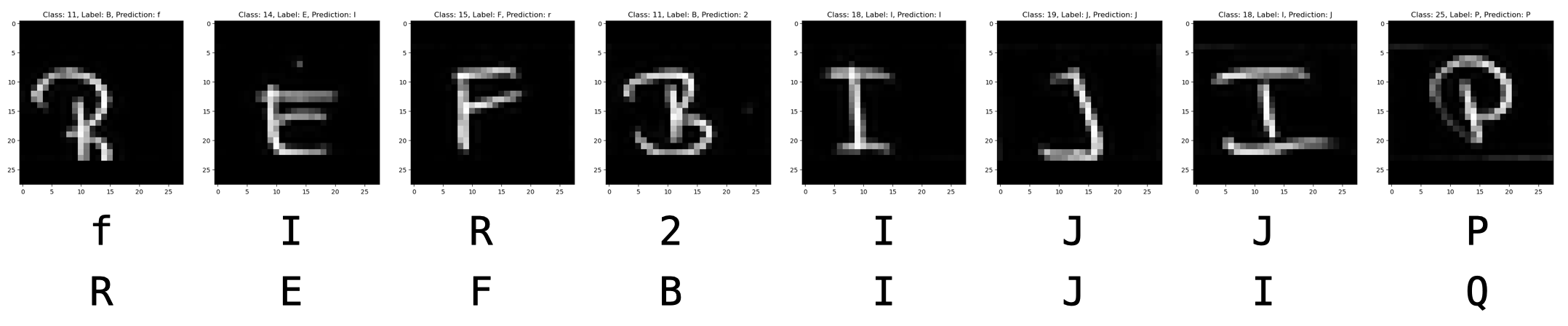}
\caption{Eight letters cropped from the table and their predicted classes.
First line shows the prediction without data augmentation, second line with
data augmentation.}
\label{fig:croppred}
\end{figure}

Using the proposed approach above for classification of the letters in our
exam, the CNN achieved an accuracy of only \SI{44.46}{\percent} (5 epochs, no
data augmentation). However, comparing the EMNIST-letters
(Fig.~\ref{fig:emnistpred}) and the cropped letters (Fig.~\ref{fig:croppred})
show, that the cropped letters are smaller since the writers usually don't
exploit the full cell size. In order to improve the cropping, a region of
interest (ROI) extraction approach similar to the one used for the
EMNIST-Letterset~\cite{r36} could be used, but due to artefacts derived from
the cell borders and accidentally drawn dots and dashes this approach is
error-prone. However, it turned out that this effect can easily be compensated
with data augmentation by zooming out the images of the training set with a
factor up to 3, which improved the accuracy up to \SI{81.93}{\percent}.

Further investigation revealed the following remaining problems:
\begin{itemize}
\item \textbf{Cropped letters:} Letters are not written exactly inside the
table cell, but protrude from the bottom or top of the cell (e.g.\ the first
letter `B' in Fig.~\ref{fig:croppred}).
\item \textbf{Unaligned cells:} The paper was scanned at an angle and the
cropped images contain border line artefacts from the table (e.g.\ lower side
of the last cell containing the `P' in Fig.~\ref{fig:croppred}).
\item \textbf{Misplaced letters:} From the 2646 cells of 49 exams
\SI{86}{\percent} (2285) are written inside the table cell correctly.
\SI{14}{\percent} of the letters are written outside the table, because the
user corrected himself. From these 2285 letters 1872 (\SI{81.93}{\percent})
letters are classified correctly with the CNN. However, the letters outside the
table haven't been segmented and classified.
\end{itemize}

\begin{figure}[t]
\centering
\includegraphics[width=0.6\textwidth]{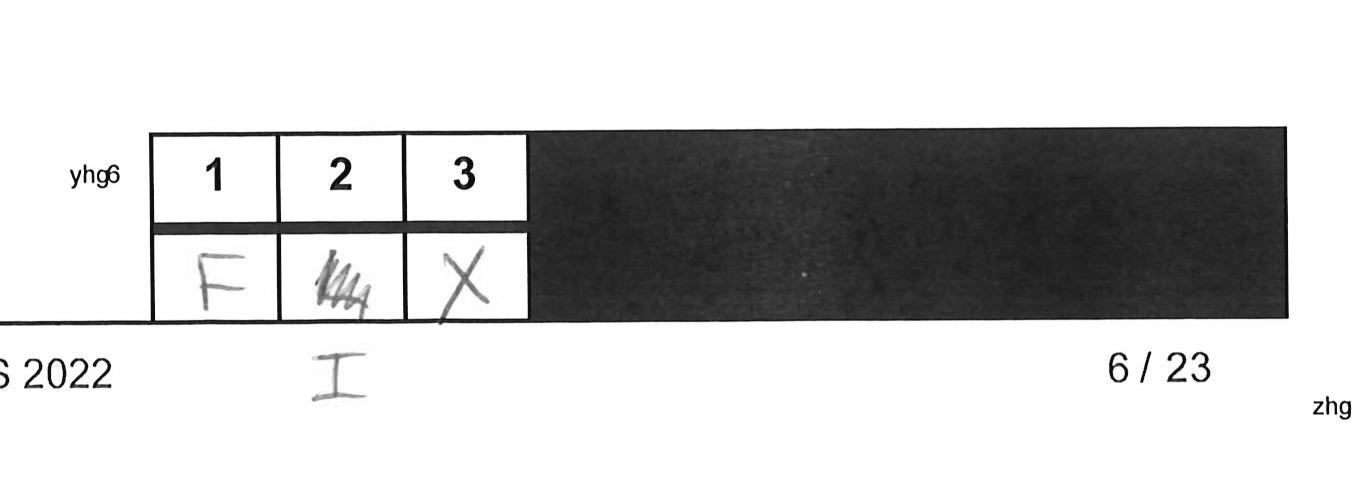}
\caption{Misplaced letter: `I' is written below the table.}
\label{fig:misplaced}
\end{figure}

The overall recognition rate incorporating the misplaced letters calculates to
\SI{70.75}{\percent}.

\section{Advanced Approach}
\subsection{Architecture}
\begin{figure}[t]
\centering
\includegraphics[width=\textwidth]{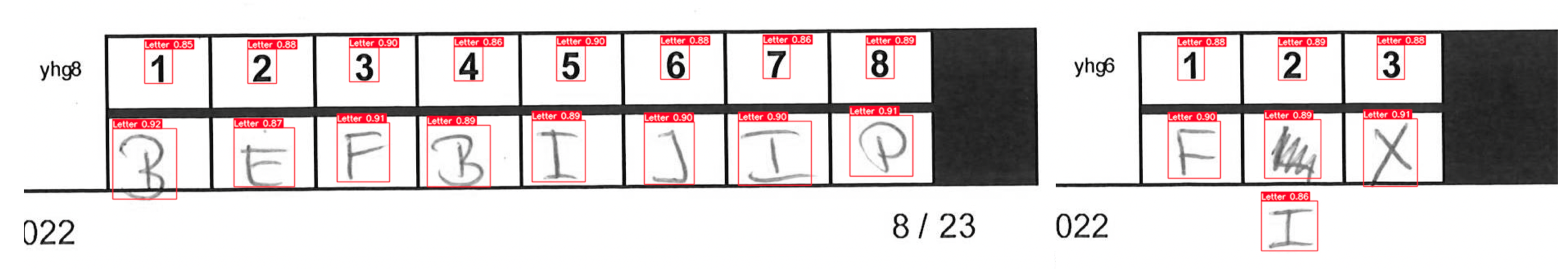}
\caption{Segmentation of digits and the letters with the YOLOv5s model. Letter
outside the table are detected, too (right).}
\label{fig:yolodetect}
\end{figure}

To overcome the segmentation problem mentioned above, a more advanced approach
has been developed, which improves the segmentation of the letters and handles
the placement of the letters above or below the table. A YOLOv5 model was used
as target detection model and trained to detect the letters in, above or below
the table. In order become independent from hardcoded cell positions and sizes,
the model was trained to detect the printed digits above the letters, too
(Fig.~\ref{fig:yolodetect}). In the Post-Processing step, the position of each
digit is determined and the letter in the corresponding column can be derived.
If the model detects more than one letter above or below the digit, the
detection with the highest probability is taken. (Since the EMNIST-Letterset
doesn't contain printed digits, for classifying the digits a separate CNN
similar to the one presented in 4.1 was trained with printed digits.)

For training the YOLOv5 model, 756 letters were marked by hand and aggregated
in on single class. A YOLOv5s (small) model with an image size of 640 pixel was
trained in 100 epochs. A mean Average Precision (mAP) with Intersection over
union (IoU) larger 0.5 was achieved in \SI{99.5}{\percent} of all samples of
the validation set (mAP\_$>$0.5 with \SI{99.5}{\percent}). The detected letters
are then cropped, resized to 28x28 shape and classified with the trained model
presented in 4.1. However, the letters cropped with the YOLOv5 model fit in size
to the training letters, so no zoom out during the data augmentation was
necessary. Letters or digits left or right outside the text markers position
(e.g.\ `yhg8' and `zhg') are ignored, if they might occur.

\subsection{Results}
\begin{figure}[t]
\centering
\includegraphics[width=\textwidth]{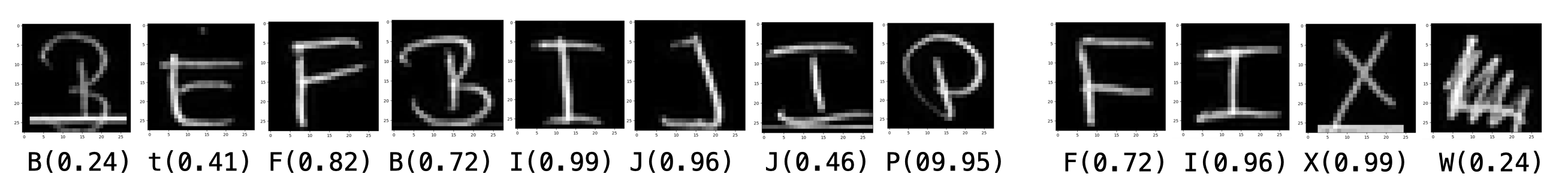}
\caption{Classification of the segmented images and their probabilities. Left:
The artifacts caused by the cropped table borders create classification issues
(`B' and `J' with low probabilities $<$0.5). Right: If two letters occur in one
column, the one with higher probability (`X' instead of `W') is taken.}
\label{fig:yoloclass}
\end{figure}

Using the YOLOv5s model for segmentation, we detected 2634 out of all 2646
(\SI{99.54}{\percent}) letters, 12 letters were not detected (e.g.\ the
corresponding digit was not found, because it was painted by the writer). From
the 2634 detected letters only 1733 (\SI{64.79}{\percent}) were classified
correctly. Fig.~\ref{fig:yoloclass} shows the cropped images and their
predictions.

Using the new approach, nearly all (\SI{99.54}{\percent}) letters are detected,
but the classification accuracy of each letter decreases from
\SI{81.93}{\percent} down to \SI{64.79}{\percent}. Further investigation
reveals, that the YOLO-network crops the letters in full size, which leads to
more artefacts inside the cropped images caused by the table borders
(Fig.~\ref{fig:yoloclass}, letter `B' and letter `I'). These lines are
represented by high values, but not part of the EMNIST training dataset, so
that their impact is high, which leads to higher rate of wrong classification.
The overall recognition rate decreases to \SI{65.50}{\percent}.

\section{Final Approach}
\subsection{Architecture}
\begin{figure}[t]
\centering
\includegraphics[width=0.7\textwidth]{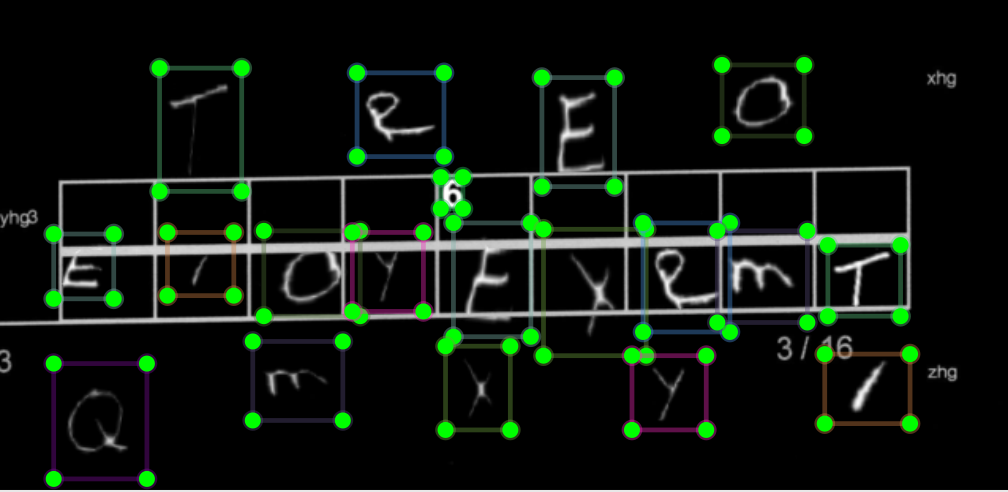}
\caption{18 letters and one digit are projected into the table cells with
random spatial deviation. The bonding boxes are calculated during the
projection.}
\label{fig:projection}
\end{figure}

To overcome the classification and segmentation problem, we generated our own
training data by using the EMNIST dataset and projected the letters into the
table cells at randomized positions. Both, the segmentation and the
classification task are executed by the YOLO-network. A variety of empty tables
were scanned and into, below and above the empty tables arbitrary letters from
the EMNIST dataset where projected. The letters (28x28 shape) were resized with
bilinear interpolation to fit the table size. To increase variety, resizing was
executed horizontally and vertically in a random way and random rotation up to
10 degree was executed. Additionally, the letters were eroded with a 3x3 kernel
in 1 or 2 iterations to make the stroke thinner. To enable the recognition of
digits, a digit was syntactically generated and projected into each sample. The
bounding box of the letters and the digit were calculated and stored in the
corresponding label file. Fig.~\ref{fig:projection} shows one sample of the
generated dataset including the calculated bounding boxes.

\subsection{Results}
\begin{figure}[t]
\centering
\includegraphics[width=\textwidth]{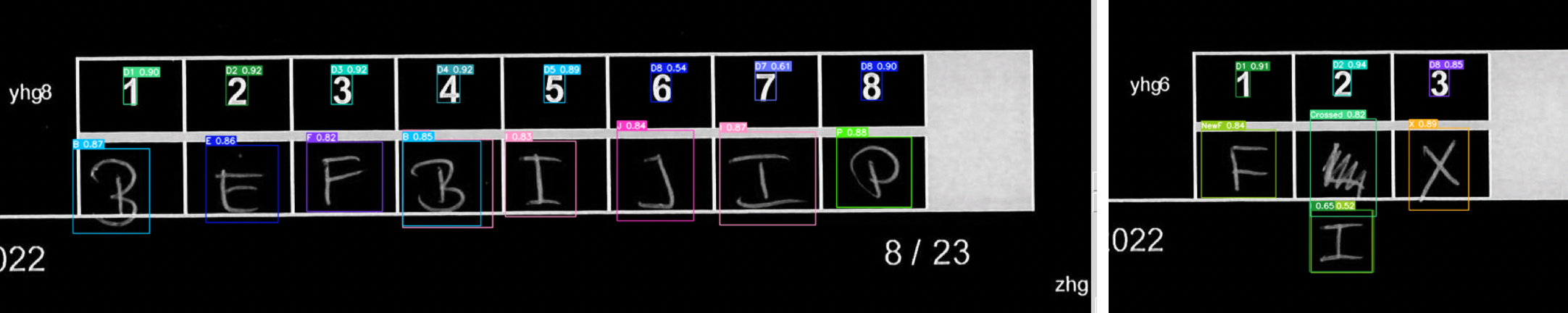}
\caption{Segmentation and classification of the letters with the trained YOLOv5
model.}
\label{fig:yolofinal}
\end{figure}

For training 12533 samples were generated and the YOLOv5s model was trained
during 100 epochs. A bounding box regression loss (box\_loss) of
\SI{1.3}{\percent}, a classification loss (class\_loss) of \SI{1.5}{\percent}
and a mean Average Precision (mAP) at IoU (Intersection over Union) threshold of
0.5 (mAP $>$0.5) of \SI{91.1}{\percent} was reached. Fig.~\ref{fig:yolofinal}
shows the classification results with the trained YOLO-model of one sample.

From 2646 letters 2631 (\SI{99.4}{\percent}) were detected, 2235
(\SI{84.95}{\percent}) of these letters were classified correctly leading to an
over-all recognition rate of \SI{84.46}{\percent}.

\subsection{EMNIST shortcoming}
\begin{figure}[t]
\centering
\includegraphics[width=\textwidth]{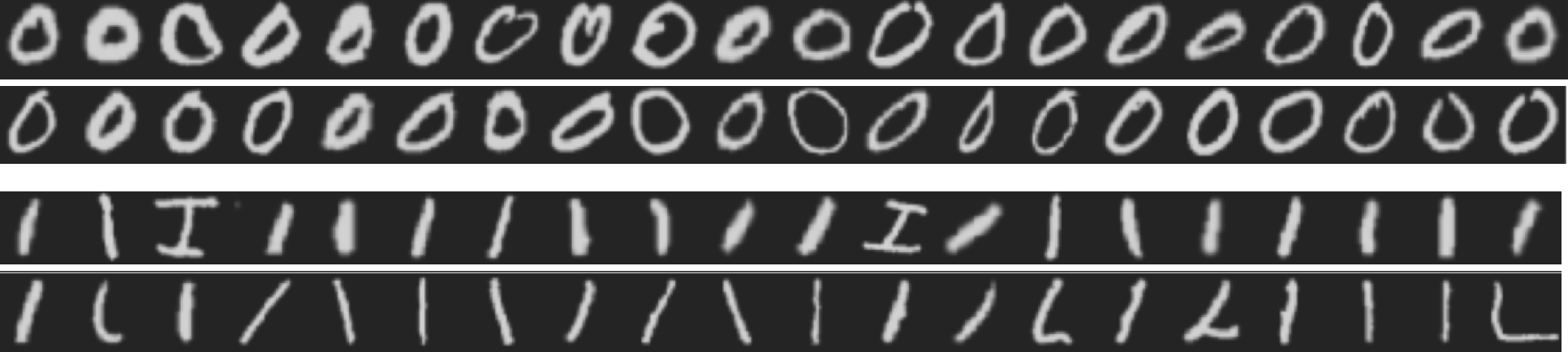}
\caption{First 20 samples of class `O' (first line), of class `0' (digit 0)
(the second line), of class `I' (third line) and of class `L' (fourth line)
from the EMNSIT Balanced Dataset.}
\label{fig:emnistshort}
\end{figure}

\begin{figure}[t]
\centering
\includegraphics[width=0.85\textwidth]{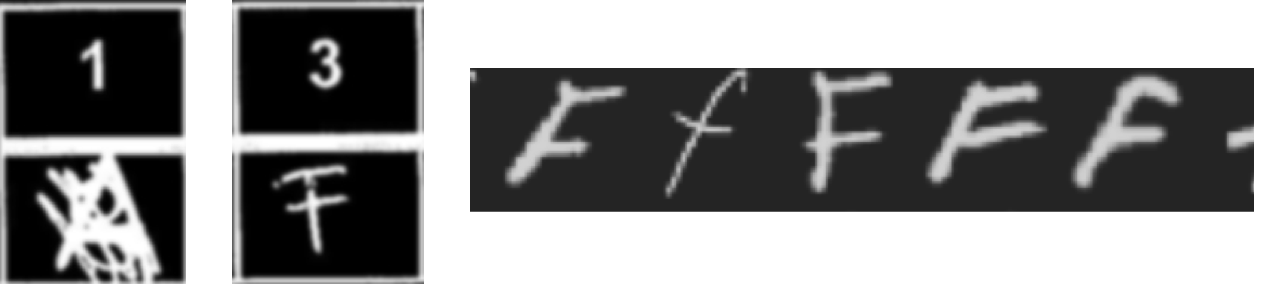}
\caption{Crossed out letters are falsely classified (left). The letter `F'
comes in two shapes (middle, right), but only one shape (right) is part of
EMNIST data set.}
\label{fig:fshapes}
\end{figure}

When the incorrectly classified letters are analyzed, the following causes are
detected:
\begin{enumerate}
\item \textbf{Letter `O' and digit `0':} Frequently instead of letter `O' the
digit `0' was recognized. It turned out that in the ``EMNSIT Balanced Letter''
dataset letter `O' and digit `0' look the same (Fig.~\ref{fig:emnistshort}, line
1 and 2).
\item \textbf{Letter `I' and `L':} Frequently, letter `I' was classified as
`L'. In the ``EMNIST Balanced Letter'' dataset letter `l' (lowercase `L') was
merged with the class of letter `L' (uppercase `L') and letter `i' (lowercase
`I') was merged with `I' (uppercase `I'). However, it turned out that by some
writers `I' (uppercase `I') and `l' (lowercase `I') look equal (just a vertical
stroke). This leads to the effect, that letter `I' is frequently classified as
`l' (lowercase `L') which belongs to the class `L' (uppercase `L').
Fig.~\ref{fig:emnistshort}, line 3 and 4 illustrates the problem.
\item \textbf{Letter `F':} Frequently, letter `F' was recognized with low
probability or even wrong classified. We observed, that some writers write the
letter `F' like a `T' with a horizontal stroke in the middle
(Fig.~\ref{fig:fshapes}, middle). However, the EMNIST data set does not contain
this shape of `F', which leads to wrong classifications.
\item \textbf{Crossed out letters:} Crossed out letters are not part of the
EMNIST data set, but they are still classified by the system. Some of these are
classified even with high probability, because no negative examples were
trained.
\end{enumerate}

\begin{figure}[t]
\centering
\includegraphics[width=\textwidth]{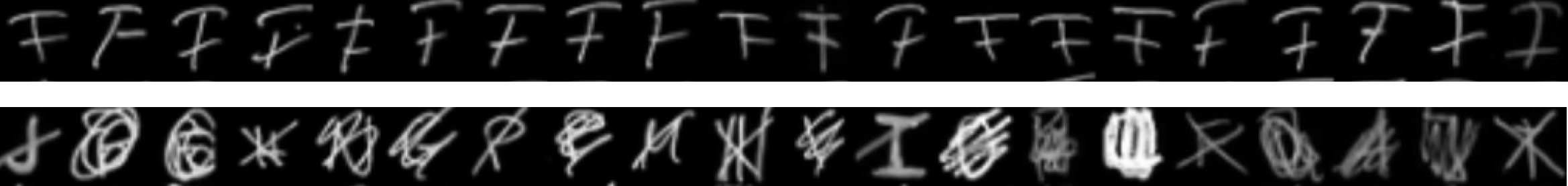}
\caption{New added classes: `F' in alternative style (first line) and crossed
out letters (second line).}
\label{fig:newclasses}
\end{figure}

To overcome these limitations, we switched to the ``EMNIST By\_Class'' dataset,
which doesn't merge small and big letters. From this dataset we extracted 2400
samples for each of the following letters only:
ABCDEFGHIJKLMNOPQRSTUVWXYZ. Numbers and small letters are omitted. Further on,
we added 240 own written samples of letter `F' in shape presented in
Fig.~\ref{fig:newclasses} (first line) in a new class called F2. Additionally,
250 samples of own generated crossed out letters were added
(Fig.~\ref{fig:newclasses}, second line).

Using the customized data set, the YOLOv5 model was trained again similar to
6.1. Based on this new trained model, from 2646 letters 2637 letters
(\SI{99.66}{\percent}) were detected, 2336 (\SI{88.59}{\percent}) of these
letters were classified correctly leading to an over-all recognition rate of
\SI{88.28}{\percent}.

\begin{table}[t]
\centering
\caption{Comparison of the different approaches.}
\label{tab:compare}
\begin{tabular}{lll}
\toprule
Used Approach & Dataset for training & Over all recognition rate \\
\midrule
Fixed cell cropping, CNN for classification & EMNIST Balanced Letter & \SI{70.75}{\percent} \\
YOLOv5 based cell cropping, CNN for classification & EMNIST Balanced Letter & \SI{65.50}{\percent} \\
YOLOv5 for segmentation and classification & EMNIST Balanced Letter & \SI{84.46}{\percent} \\
YOLOv5 for segmentation and classification & Customized dataset & \SI{88.28}{\percent} \\
\bottomrule
\end{tabular}
\end{table}

\section{Conclusion}
The current state-of-the-art methodology for recognizing handwritten characters
in forms involves the execution of segmentation, feature extraction, and
classification stages as distinct tasks. However, the use of a CNN enables the
simultaneous execution of both feature extraction and classification stages in a
single task. In this research article, we demonstrate that all three tasks,
namely segmentation, feature extraction, and classification, can be effectively
executed through a single deep neural network. The study employed the YOLOv5
model, which necessitated the synthetic generation of training data from the
underlying form and the EMNIST data set. Using this approach, the recognition
rate could be increased significantly and even letters written outside the form
could be recognized correctly. However, our analysis revealed multiple
limitations associated with the aforementioned data set. As a result, we
developed a customized version and demonstrated its efficacy in improving the
recognition rate up to \SI{88.28}{\percent}.

Obviously, the recognition rate is not high enough to execute fully automated
exam correction. To overcome this limitation, we developed an interactive system
that superimposes proposed letters on the corresponding bitmap for subsequent
human verification. In cases where incorrect corrections are made, a human
operator is required to manually enter the accurate letter. Despite the need for
human intervention, this system considerably reduces the time required for
correction by a factor of approximately five. Additionally, to ensure further
accuracy and provide a means for self-verification, the bitmaps along with the
final determined letters are transmitted to the examinee via email.

\nocite{*}
\bibliographystyle{splncs04}
\bibliography{refs}

\end{document}